%% file: main.tex
\documentclass[runningheads]{llncs}
\usepackage[T1]{fontenc}

\usepackage[dvipdfmx]{graphicx} 
\usepackage{hyperref}
\usepackage{url}

\usepackage{algorithm}
\usepackage{algpseudocode}
\input{math_commands}

\usepackage{cleveref}
\usepackage{booktabs}
\usepackage{multirow}
\usepackage{wrapfig}
\usepackage{subcaption}
\usepackage{threeparttable}
\usepackage{color}
\usepackage[numbers]{natbib}

\newcommand{\iter}[2]{#1^{(#2)}}

\begin{document}
\title{Enhancing Output Diversity Improves Conjugate Gradient-based Adversarial Attacks}
\titlerunning{Enhancing Output Diversity}
%
\author{Keiichiro Yamamura\inst{1}\orcidID{0000-0003-4696-2881} \and
Issa Oe\inst{1}\orcidID{0009-0001-7216-2143} \and
Hiroki Ishikura\inst{1}\orcidID{0000-0002-4979-5276} \and
Katsuki Fujisawa\inst{2}\orcidID{0000-0001-8549-641X}}
\authorrunning{K. Yamamura et al.}
%
\institute
{Graduate School of Mathematics, Kyushu University, Fukuoka, Japan.
\and
Institute of Mathematics for Industry, Kyushu University, Fukuoka, Japan.
\email{keiichiro.yamamura@kyudai.jp}\\
}

\maketitle
\begin{abstract}
Deep neural networks are vulnerable to adversarial examples, and adversarial attacks that generate adversarial examples have been studied in this context.
Existing studies imply that increasing the diversity of model outputs contributes to improving the attack performance.
This study focuses on the Auto Conjugate Gradient (ACG) attack, which is inspired by the conjugate gradient method and has a high diversification performance. We hypothesized that increasing the distance between two consecutive search points would enhance the output diversity.
To test our hypothesis, we propose Rescaling-ACG (ReACG), which automatically modifies the two components that significantly affect the distance between two consecutive search points, including the search direction and step size.
ReACG showed higher attack performance than that of ACG, and is particularly effective for ImageNet models with several classification classes.
Experimental results show that the distance between two consecutive search points enhances the output diversity and may help develop new potent attacks. The code is available at \url{https://github.com/yamamura-k/ReACG}.
\end{abstract}

\section{Introduction}
Szegedy et al.\cite{szegedy2013intriguing} noted that the output of deep neural networks (DNNs) can be significantly altered by small perturbations imperceptible to the human eye. These perturbed inputs are referred to as adversarial examples, and it is well-known that this is a crucial vulnerability of DNNs.
Addressing this vulnerability is crucial because DNNs have safety-critical applications such as automated driving \cite{gupta2021deep}, facial recognition \cite{adjabi2020past}, and cyber security \cite{liu2022inferring}.
Several defense mechanisms have been proposed to improve the robustness of DNNs against adversarial examples. The most fundamental approach is adversarial training~\cite{madry2018towards}, which requires several adversarial examples. 
The rapid generation of adversarial examples is expected to reduce the computation time for adversarial training and improve the robustness of the obtained DNN.

Adversarial attacks are methods used to generate adversarial examples.
If the purpose is better robustness evaluation and improving robustness, the white-box attack, which optimizes the objective function $L$ using the gradient of the DNN to generate adversarial examples, is one of the most promising approaches.
Untargeted attacks, in which attackers do not specify a misclassification class, are often used for robustness evaluations and adversarial training. This study focuses on untargeted white-box attacks against image classifiers.

Targeted attacks have also been studied \cite{8424625,Li_2020_CVPR,yao2020miss}, in which attackers specify the misclassification class, in contrast to untargeted attacks.
Different adversarial examples with different predictions can be generated through targeted attacks on a single input.
The difference in their predicted classes implies differences in the model outputs. 
The success of existing untargeted attacks that consider output diversity \cite{MTPGD,tashiro2020ods,Yamamura2022} suggests that enhancing the output diversity results in higher attack performance.

We focus on the Auto Conjugate Gradient (ACG) attack \cite{Yamamura2022}, which is based on the conjugate gradient method and has a high diversification performance.
ACG does not assume any restarts, and is advantageous over restart-based approaches in terms of the computation time \cite{MTPGD,tashiro2020ods}.
Restart-based techniques can also be combined with ACG.
The search points of ACG move more in each iteration than those of Auto-PGD (APGD)~\cite{croce2020reliable}, which is based on the momentum method. Additionally, ACG shows a higher diversity of second likely predictions (CW target class, CTC) than that in APGD.
We hypothesized that increasing the distance between two consecutive search points would enhance CTC diversity.

To validate this hypothesis, we propose Rescaling-ACG (ReACG), which modifies the search direction and step size control of ACG, as described in \cref{sec:ascg}.
The search direction and step size significantly affected the distance between two consecutive search points.
ReACG determines whether the coefficient $\iter{\beta}{k}$ should be modified based on the ratio of the gradient to the conjugate gradient (\cref{sec:rescaling_condition}).
If $\iter{\beta}{k}$ must be modified, $\iter{\beta}{k}$ is calculated using the gradient normalized by its Euclidean norm. 
Although ACG has different characteristics that those of APGD, it updates the step size using the same rule as that in APGD.
By contrast, as described in \cref{sec:stepsize_optimization,sec:experiments_tuning}, ReACG controls the step size with appropriate checkpoints determined through multi-objective optimization using Optuna \cite{optuna}.

\Cref{sec:experiments_main} presents the empirical evaluation of ReACG on 30 robust models trained on the CIFAR-10, CIFAR-100 \cite{Krizhevsky09learningmultiple}, and ImageNet \cite{ILSVRC15} datasets.
ReACG showed a higher attack performance than that of the state-of-the-art (SOTA) methods APGD and ACG for more than 90\% of the 30 models, including a wide range of architectures. Particularly, ReACG exhibited the highest attack performance for all models trained on ImageNet, which has several classification classes.
ReACG showed a higher attack performance than those of APGD and ACG, with approximately 0.4 to 0.9\% and 0.1 to 0.4\%, respectively.
Considering the recent advances in adversarial attacks based on nonlinear optimization methods, such as PGD~\cite{madry2018towards}, APGD, and ACG, significant performance improvement was achieved.

The analyses in \cref{sec:analysis_rescaling,sec:analysis_ctc_variation} show that modifying $\iter{\beta}{k}$ and step size control increases the distance between two consecutive search points and the CTC diversity, which leads to the high attack performance of ReACG. 
This relationship is beneficial for the development of novel and potent attacks.
The contributions of this study are summarized below. \\
1. We propose ReACG, which automatically modifies $\iter{\beta}{k}$ based on the hypothesis that increasing the distance between two consecutive search points enhances output diversity. \\
2. ReACG showed a higher attack performance than those of SOTA attacks on more than 90\% of the 30 robust models trained on three representative datasets. \\
3. Our analysis suggests that increasing the distance between two consecutive search points enhances CTC diversity. This relationship is beneficial for the development of novel and potent attacks.

\section{Preliminaries}
\subsection{Problem setting}
Let $f: D \to \mathbb{R}^C$ be a locally differentiable function that serves as a $C$-classifier. Assume that the input $\bm{x}_{\text{org}}$ is classified into class $c$ using classifier $f$. Given a positive number $\varepsilon$ and distance function $d: D \times D \to \mathbb{R}$, we define an adversarial example $\bm{x}_{\text{adv}} \in D$ as an input satisfying the following conditions
\begin{align}
    \arg\max_{i} f_{i}(\bm{x}_{\text{adv}}) \neq c, \quad d(\bm{x}_{\text{org}}, \bm{x}_{\text{adv}}) \leq \varepsilon.
\end{align}
Generally, an adversarial example is generated by maximizing the objective function $L$ within the feasible region $S = \{\bm{x} \in D \mid d(\bm{x}_{\text{org}}, \bm{x}_{\text{adv}}) \leq \varepsilon\}$. 
This problem is formulated as follows: 
\begin{align}
    \label{eq:problem_maximization} \max_{\bm{x} \in S} L(f(\bm{x}), c)
\end{align}
The condition $\arg\max_{i} f_{i}(\bm{x}_{\text{adv}}) \neq c$ can be rephrased as $\max_{i\neq c} f_i(\bm{x}_{\text{adv}}) - f_c(\bm{x}_{\text{adv}}) \geq 0$. Hence, objective functions such as the CW loss ($L_\textrm{CW}$)~\cite{CW2017} and Difference of Logit Ratio (DLR) loss ($L_\textrm{DLR}$)~\cite{croce2020reliable} are commonly used.
Let $\bm{z}=f(\bm{x})$ and $\pi_k$ be the index of the k-th largest element of $\bm{z}$.
The CW loss and DLR loss are denoted as $L_\textrm{CW}(\bm{z}, c)=-\bm{z}_c + \max_{i\neq c}\bm{z}_i$ and $L_\textrm{DLR}(\bm{z}, c)=\frac{L_\textrm{CW}(\bm{z}, c)}{\bm{z}_{\pi_1} - \bm{z}_{\pi_3}}$, respectively.
$\arg\max_{i\neq c}f_i(\bm{x})$ was referred to as \emph{CTC} in \cite{Yamamura2022}. We define output diversity as the variation of CTC during the attack procedure.

For adversarial attacks on image classifiers, the input space is $D = [0, 1]^n$. Additionally, the Euclidean distance $\|\cdot\|_2$ or uniform distance $\|\cdot\|_\infty$ is often used as the distance function $d$.
This study focuses primarily on $\ell_\infty$ attacks that use $d(\bm{u}, \bm{v}) = \|\bm{u} - \bm{v}\|_\infty$ as a distance function.

\subsection{Related work}
Szegedy et al. \cite{szegedy2013intriguing} suggested the existence of adversarial examples, and several gradient-based attacks have been proposed subsequently \cite{CW2017,croce2020reliable,CGD2022,madry2018towards,Yamamura2022}. Among these, a promising approach is to consider the diversity in the output space. MT-PGD \cite{MTPGD} achieves diversification in the output space by sequentially performing targeted attacks using different misclassified target classes. Output Diversified Sampling \cite{tashiro2020ods} and its variants \cite{Ye2022Practical} randomly sample the initial point, considering the diversity in the output space by maximizing the inner product of a random vector $\bm{w}$ and logit $f(\bm{x})$.
These methods assume several restarts based on the changes in the target class or initial points.
By contrast, ACG improves the output diversity by moving to the sign of the conjugate gradient. 
ACG is advantageous in terms of the computational time because output space diversification can be achieved without restarts.
Additionally, an ensemble of attacks, such as AutoAttack \cite{croce2020reliable}, has attracted research attention in recent years ~\cite{AutoAE23,CAA2021}.
Although ensemble-based attacks are stronger than simple attacks, such as PGD, their performance depends on the individual attacks in the ensemble.
Therefore, it is important to develop individualized attacks.

\subsection{Auto Conjugate Gradient attack}
\subsubsection{ACG step}
ACG attack~\cite{Yamamura2022} is based on the conjugate gradient method, which updates search points using the following formulas:
\begin{align}
    \label{eq:grad}\iter{\bm{g}}{k} &\gets \nabla L\left(f(\iter{\bm{x}}{k}), c\right), \\
    \label{eq:grad_diff}\iter{\bm{y}}{k-1} &\gets \iter{\bm{g}}{k-1} - \iter{\bm{g}}{k}, \\
    \label{eq:beta_hs}\iter{\beta}{k}_{HS} &\gets -\frac{\langle-\iter{\bm{g}}{k},\iter{\bm{y}}{k-1}\rangle}{\langle\iter{\bm{s}}{k-1},\iter{\bm{y}}{k-1}\rangle}, \\
    \label{eq:conj_grad}\iter{\bm{s}}{k}&\gets\iter{\bm{g}}{k} + \iter{\beta}{k}_{HS}\iter{\bm{s}}{k-1}, \\
    \label{eq:update_xk}\iter{\bm{x}}{k+1} &\gets P_{S}\left(\iter{\bm{x}}{k}+\iter{\eta}{k}\mathrm{sign}(\iter{\bm{s}}{k})\right),
\end{align}
where $\langle\cdot,\cdot\rangle$ is the inner product and $P_S(\cdot)$ is a projection onto the feasible region $S$. $\iter{\bm{x}}{k}$ is referred to as the search point, and $\iter{\bm{s}}{k}$ is referred to as search direction.

\subsubsection{Step-size updating rule}
ACG controls the step size using the same rule as that in APGD.
APGD halves the step size and moves to the incumbent solution $\bm{x}_{adv}$ if conditions C1 and C2 are satisfied at the precalculated checkpoints $w_j\in W$.
\begin{enumerate}
    \item[C1] $\displaystyle\sum_{i=w_{j-1}}^{w_j - 1}\bm{1}_{L\left(f(\bm{x}^{(i+1)}), c\right) > L\left(f(\bm{x}^{(i)}), c\right)}<\rho\cdot(w_j - w_{j-1})$ \label{eq:condition1}
    \item[C2] $\displaystyle L_{\textrm{max}}\left(f(\bm{x}^{(w_{j-1})}), c\right)=L_{\textrm{max}}\left(f(\bm{x}^{(w_{j})}), c\right)$ and $\displaystyle\eta^{(w_{j-1})} = \eta^{(w_{j})}$, \label{eq:condition2}
\end{enumerate}
where $\rho\in[0, 1]$ denotes a parameter with a default value of $0.75$.
The following gradual equation determines the checkpoints $w_j$:
\begin{align}
    &p_0=0,~ p_1=0.22, ~q=0.03, ~q_{min}=0.06 \\
    &p_{j+1}=p_{j} + \max(p_{j}-p_{j-1}-q, q_{min}),~w_{j}=\lceil p_{j}N\rceil.
\end{align}

\section{Rescaling-ACG}
\label{sec:ascg}
This section describes the ReACG attack, a modification of ACG aimed at increasing the distance between two consecutive search points and enhancing output diversity.
ReACG automatically modifies its search direction and controls the step size using the appropriate checkpoints obtained through multi-objective optimization.
\Cref{sec:ascg_direction} describes the modification of the search direction, and \cref{sec:stepsize_optimization} provides the checkpoints used for step size calculation.
The pseudocode for ReACG is described in \cref{algo:ascg}.
\subsection{Search direction}
\label{sec:ascg_direction}
\subsubsection{Motivation}
\label{sec:motivation}
Search direction such as $\iter{\bm{s}}{k}$ or $\iter{\bm{g}}{k}$ is one of the main factors affecting the distance between two consecutive search points. Conjugate gradient methods are characterized by the coefficient $\iter{\beta}{k}$ used to calculate the conjugate gradient. The preliminary experiment conducted by Yamamura et al. \cite{Yamamura2022} suggests that $\iter{\beta}{k}$ significantly affects attack performance. 
\begin{figure}[t]
    \centering
    \includegraphics[width=0.85\linewidth]{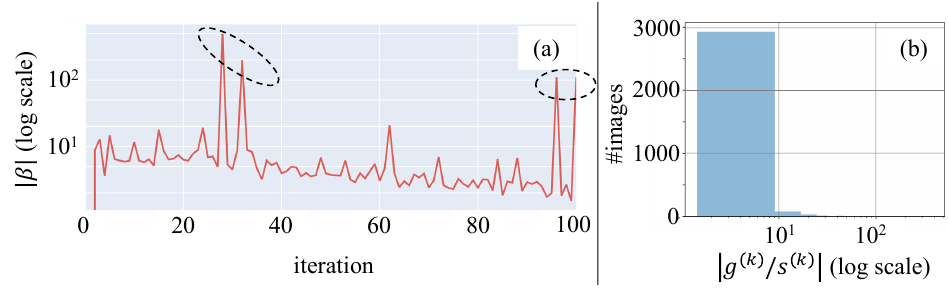}
    \caption{(a) Transition of $|\iter{\beta_{HS}}{k}|$.
    (b) Distribution of 
    $|\iter{\bm{g}}{k}/\iter{\bm{s}}{k}|$.
    }
    \label{fig:beta_transition}
\end{figure}
\Cref{fig:beta_transition} (a) depicts the average transition of $|\iter{\beta}{k}_{HS}|$ over 10,000 images. Although $|\iter{\beta}{k}_{HS}|$ was less than 10 in several iterations, it reached approximately 500 in other iterations.
Additionally, \cref{fig:beta_transition} (b) shows that most elements of $|\iter{\bm{g}}{k}/\iter{\bm{s}}{k}|$ averaged over 10,000 images are less than 10.
\Eqref{eq:conj_grad} indicates that if $|\iter{\beta}{k}_{HS}|$ is extremely larger than each element of $|\iter{\bm{g}}{k}/\iter{\bm{s}}{k}|$, $\iter{\bm{s}}{k+1}$ and $\iter{\bm{s}}{k}$ are likely to be the same vector.
These results imply that an ACG search may be redundant.
\begin{algorithm}
    \centering
    \caption{ReACG}\label{algo:ascg}
    \begin{algorithmic}[1]
        \Require{$f$, $L$, $S$, $c$, $\iter{\bm{x}}{0}$, $\iter{\eta}{0}$, $N$, $W=\{w_0,\ldots,w_n\}$}
        \Ensure{$\bm{x}_{adv}$}
        \State{$\bm{x}_{adv}\gets\iter{\bm{x}}{0};\iter{\beta}{0}\gets 0;\iter{\bm{s}}{0}\gets\nabla L(f(\iter{\bm{x}}{0}), c);\bm{x}_{pre}\gets\iter{\bm{x}}{0};\bm{s}_{pre}\gets\iter{\bm{s}}{0}$}
        \For{$k=0$ to $N$}
            \State{Compute $\iter{\bm{x}}{k+1}$ (\ref{eq:update_xk}); $\iter{\eta}{k+1}\gets\iter{\eta}{k}$}
            \If{$L(f(\iter{\bm{x}}{k+1}), c) > L(f(\bm{x}_{adv}), c)$}
                \State{$\bm{x}_{adv}\gets\iter{\bm{x}}{k+1};\bm{x}_{pre}\gets\iter{\bm{x}}{k};\bm{s}_{pre}\gets\iter{\bm{s}}{k}$}
            \EndIf
            \If{$k\in W$ and (C1 or C2 hold)}
                \State{$\iter{\eta}{k+1}\gets\iter{\eta}{k}/2;\iter{\bm{x}}{k+1}\gets\bm{x}_{adv};\iter{\bm{x}}{k}\gets\bm{x}_{pre};\iter{\bm{s}}{k}\gets\bm{s}_{pre}$}
            \EndIf
            \State{/* The difference to ACG is highlighted in \textcolor{blue}{blue}. */}
            \State{Compute $\iter{\beta}{k+1}$ (\ref{eq:beta_hs}) and \textcolor{blue}{$\iter{\tilde{\beta}}{k}$ (\ref{eq:beta_hs_tilde})}}
            \If{\textcolor{blue}{$|\iter{\tilde{\beta}}{k}| < |\iter{\beta}{k}|$ and (\ref{eq:beta_condition}) hold}}
                \State{\textcolor{blue}{$\iter{\beta}{k}\gets\iter{\tilde{\beta}}{k}$}}
            \EndIf
            \State{Compute $\iter{\bm{s}}{k+1}$ (\ref{eq:conj_grad})}
        \EndFor
    \end{algorithmic}
\end{algorithm}

\subsubsection{Rescaling condition}
\label{sec:rescaling_condition}
\eqref{eq:conj_grad} suggests that $\iter{\bm{s}}{k}_i$ and $\iter{\bm{s}}{k-1}_i$ are likely to take appximately the same values for the index $i$ such that $\left|\iter{\bm{g}}{k}/\iter{\bm{s}}{k-1}\right|_i \ll |\iter{\beta}{k}|$.
If the following inequality holds, ReACG modifies the coefficient $\iter{\beta}{k}$:
\begin{equation}
    \label{eq:beta_condition}\mathbb{E}\left[\left|\frac{\iter{\bm{g}}{k}}{\iter{\bm{s}}{k-1}}\right|\right]<|\iter{\beta}{k}|
\end{equation}
The severity of this condition can be adjusted by multiplying the right side of \eqref{eq:beta_condition} by a constant. However, we did not consider this extension to maintain a number of hyperparameters similar to those of ACG.

\subsubsection{Rescaling method}
\label{sec:rescaling_method}
When \eqref{eq:beta_condition} holds, the coefficient $\iter{\tilde{\beta}}{k}_{HS}$ is calculated using $\iter{\bm{\tilde{g}}}{k}=\iter{\bm{g}}{k}/\|\iter{\bm{g}}{k}\|_2$ instead of $\iter{\bm{g}}{k}$.
More precisely, equations \ref{eq:y_tilde} and \ref{eq:beta_hs_tilde} are used to calculate the coefficients instead of equations \ref{eq:grad_diff} and \ref{eq:beta_hs}.
\begin{align}
    \iter{\bm{y}}{k-1} & \gets\iter{\bm{\tilde{g}}}{k-1} - \iter{\bm{\tilde{g}}}{k} \label{eq:y_tilde}\\
    \iter{\tilde{\beta}}{k}_{HS} & \gets -\frac{\langle-\iter{\bm{\tilde{g}}}{k}, \iter{\bm{\tilde{y}}}{k-1}\rangle}{\langle\iter{\bm{s}}{k-1}, \iter{\bm{\tilde{y}}}{k-1}\rangle}\label{eq:beta_hs_tilde}
\end{align}
If $|\iter{\tilde{\beta}}{k}_{HS}| < |\iter{\beta}{k}_{HS}|$ holds, $\iter{\bm{s}}{k}$ is calculated using $\iter{\tilde{\beta}}{k}_{HS}$ in stead of $\iter{\beta}{k}$.

\subsection{Rethinking the step size}
\label{sec:stepsize_optimization}
ACG uses precalculated checkpoints to control the step size. The three magic numbers $p_1,q$, and $q_{min}$ that appear in the checkpoint calculation affect the obtained checkpoints. We searched for the appropriate values of these parameters through multi-objective optimization using Optuna. Based on the experimental results in \cref{sec:experiments_tuning}, we set $p_1=0.43, q=0.24$, and $q_{min}=0.08$.

\section{Experiments}
\label{sec:experiments}
This section describes the results of comparative experiments on $\ell_\infty$-robust models trained on CIFAR-10, CIFAR-100, and ImageNet.
The attacked models were in RobustBench \cite{croce2021robustbench}, which is a well-known benchmark for adversarial robustness.
The performance evaluation was based on robust accuracy, defined as follows:
\begin{equation}
    \frac{\textrm{\# correctly classified adversarial examples}}{\textrm{\# test samples}}\times 100.
\end{equation}
A low robust accuracy indicates a high misclassification rate and high attack performance.
As with the RobustBench leaderboard, we used 10,000 images and $\varepsilon=8/255$ for CIFAR-10/100, and 5,000 images and $\varepsilon=4/255$ for ImageNet.
The experiments were conducted using an Intel(R) Xeon(R) Silver 4214R CPU at 2.40GHz and NVIDIA RTX A6000.
We chose APGD and ACG as the baseline. Both typically perform better for adversarially robust models than early techniques such as FGSM \cite{goodfellow2014explaining} and PGD \cite{madry2018towards}.

\subsection{Experiments on step size control}
\label{sec:experiments_tuning}
\begin{table}[tb]
    \centering
    \caption{The results of step size optimization using Optuna. The lowest robust accuracy for each model and loss is in bold, and the second lowest is underlined.}
    \label{tab:tuning}
    \scalebox{0.83}{
    \begin{tabular}{ccc ccccc ccccc}
    \toprule
        \multicolumn{3}{c}{Loss} & \multicolumn{5}{c}{CW} & \multicolumn{5}{c}{DLR} \\
        \cmidrule(lr){1-3}
        \cmidrule(lr){4-8}
        \cmidrule(lr){9-13}
        $p_1$ & $q$ & $q_{min}$ & \cite{Sehwag2021Proxy} & \cite{Wong2020Fast}& \cite{Addepalli2022Efficient} & \cite{Andriushchenko2020Understanding} & \cite{Gowal2021Improving} & \cite{Sehwag2021Proxy} & \cite{Wong2020Fast}& \cite{Addepalli2022Efficient} & \cite{Andriushchenko2020Understanding} & \cite{Gowal2021Improving} \\
        \midrule
        0.22 & 0.03 & 0.06 & 56.47 & 45.07 & 53.08 & 45.47 & 59.73 & 56.34 & 45.04 & 52.92 & 45.28 & 59.67 \\
        \textbf{0.43} & \textbf{0.24} & \textbf{0.08} & 56.48 & \textbf{45.00} & \underline{53.06} & \textbf{45.31} & \textbf{59.65} & \textbf{56.23} & 44.83 & \textbf{52.89} & \underline{45.10} & 59.47 \\
        0.26 & 0.04 & 0.08 & \textbf{56.41} & 45.07 & 53.08 & 45.43 & 59.73 & 56.38 & 45.01 & 52.92 & 45.21 & 59.58 \\
        0.31 & 0.28 & 0.09 & 56.47 & 45.06 & 53.12 & 45.41 & 59.71 & \underline{56.26} & 44.91 & \underline{52.91} & 45.25 & 59.62 \\
        0.42 & 0.25 & 0.10 & \underline{56.44} & \underline{45.01} & \textbf{53.05} & \underline{45.32} & 59.66 & \textbf{56.23} & \underline{44.82} & 52.94 & 45.12 & \underline{59.44} \\
        0.69 & 0.69 & 0.05 & 56.57 & 45.19 & 53.19 & 45.35 & \underline{59.66} & 56.27 & \textbf{44.81} & 52.96 & \textbf{45.06} & \textbf{59.41} \\
    \bottomrule
    \end{tabular}
    }
\end{table}
    This experiment used five robust PreActResNets \cite{Addepalli2022Efficient,Andriushchenko2020Understanding,Gowal2021Improving,Sehwag2021Proxy,Wong2020Fast} trained on CIFAR-10.
Optuna \cite{optuna} searched for appropriate parameters through multi-objective optimization, which minimized robust accuracy and maximized the average CW loss value.
In each optimization iteration with Optuna, ReACG with CW loss was executed starting at the input points for different values of $0.01\leq p_1\leq 0.9$, $0.01\leq q \leq 0.5 p_1$, and $0.01\leq q_{min}\leq 0.1$. 
To make the step size sufficiently small, the objective function of this multi-objective optimization problem was designed to return a tuple $(\infty, -\infty)$ when the number of checkpoints was less than four. Optuna requires 100 iterations per model. \Cref{tab:tuning} shows the obtained parameters and robust accuracy. The obtained $p_1$ values tend to be larger than those in the APGD setting. The experimental results indicate that $p_1=0.43, q=0.24$, and $q_{min}=0.08$ are effective in several models.

\begin{table}
    \centering
    \caption{Comparison in robust accuracy. $\Delta$ is the difference in robust accuracy between ACG and ReACG. The lowest value for each model and loss is in bold.}
    \label{tab:full_cifar10_100}
    \scalebox{0.83}{
    \begin{tabular}{c c c cccc cccc}
    \toprule
    \multicolumn{3}{c}{CIFAR-10 ($\varepsilon=8/255$)} & \multicolumn{4}{c}{CW loss} & \multicolumn{4}{c}{DLR loss} \\
    \cmidrule(lr){1-3}
    \cmidrule(lr){4-7}
    \cmidrule(lr){8-11}
    Defense & Model & clean & APGD & ACG & ReACG & $\Delta ~(\uparrow)$ & APGD & ACG & ReACG & $\Delta ~(\uparrow)$ \\
    \midrule 
    Peng~\cite{Peng2023Robust} & WRN70-16 & 93.27 & 71.92 & 71.66 & \textbf{71.60} & \textbf{0.06} & 71.99 & 71.53 & \textbf{71.41} & \textbf{0.12} \\
    Wang~\cite{Wang2023Better} & WRN70-16 & 93.25 & 71.57 & 71.25 & \textbf{71.17} & \textbf{0.08} & 71.58 & 71.32 & \textbf{71.13} & \textbf{0.19} \\
     \addlinespace
    \multirow{2}{*}{Bai \cite{Bai2023Improving}} & RN152+ & \multirow{2}{*}{95.23} & \multirow{2}{*}{\textbf{69.42}} & \multirow{2}{*}{69.78} & \multirow{2}{*}{70.10} & \multirow{2}{*}{-0.32} & \multirow{2}{*}{\textbf{69.38}} & \multirow{2}{*}{69.86} & \multirow{2}{*}{69.72} & \multirow{2}{*}{\textbf{0.14}} \\
     & WRN70-16 &  &  &  &  &  &  &  &  &  \\
     \addlinespace
    Wang \cite{Wang2023Better}& WRN28-10 & 92.44 & 68.28 & 68.02 & \textbf{67.77} & \textbf{0.25} & 68.30 & 67.97 & \textbf{67.80} & \textbf{0.17} \\
    Rebuffi \cite{Rebuffi2021Fixing}& WRN70-16 & 92.23 & 67.71 & 67.54 & \textbf{67.45} & \textbf{0.09} & 67.85 & 67.38 & \textbf{67.24} & \textbf{0.14} \\
    Cui \cite{Cui2023Decoupled}& WRN28-10 & 92.17 & 68.57 & 68.34 & \textbf{68.24} & \textbf{0.10} & 68.63 & 68.21 & \textbf{68.06} & \textbf{0.15} \\
    Huang \cite{Huang2022Revisiting}& WRN-A4 & 91.58 & 66.79 & 66.52 & \textbf{66.41} & \textbf{0.11} & 67.02 & 66.52 & \textbf{66.34} & \textbf{0.18} \\
    Gowal20 \cite{Gowal2020Uncovering}& WRN70-16 & 91.10 & 66.76 & 66.48 & \textbf{66.47} & \textbf{0.01} & 66.89 & 66.40 & \textbf{66.31} & \textbf{0.09} \\
    Gowal21 \cite{Gowal2021Improving}& WRN70-16 & 88.74 & 67.75 & 67.26 & \textbf{67.19} & \textbf{0.07} & 68.33 & 67.21 & \textbf{67.00} & \textbf{0.21} \\
    Rebuffi \cite{Rebuffi2021Fixing}& WRN106 & 88.50 & 65.57 & \textbf{65.35} & 65.38 & -0.03 & 65.60 & 65.26 & \textbf{65.07} & \textbf{0.19} \\
    \addlinespace
    \cmidrule(lr){1-3}
    \multicolumn{3}{c}{CIFAR-100 ($\varepsilon=8/255$)} &&&&&&&&\\
    \midrule
    \multirow{2}{*}{Bai \cite{Bai2023Improving}} & RN152+ & \multirow{2}{*}{85.19} & \multirow{2}{*}{\textbf{40.70}} & \multirow{2}{*}{41.64} & \multirow{2}{*}{41.60} & \multirow{2}{*}{\textbf{0.04}} & \multirow{2}{*}{\textbf{40.78}} & \multirow{2}{*}{41.60} & \multirow{2}{*}{41.66} & \multirow{2}{*}{-0.06} \\
    & WRN70-16 &  &  &  &  &  &  &  &  &  \\
    \addlinespace
    \multirow{2}{*}{Bai \cite{Bai2023Improving}} & RN152+ & \multirow{2}{*}{80.20} & \multirow{2}{*}{\textbf{37.47}} & \multirow{2}{*}{37.66} & \multirow{2}{*}{37.48} & \multirow{2}{*}{\textbf{0.18}} & \multirow{2}{*}{37.82} & \multirow{2}{*}{37.81} & \multirow{2}{*}{\textbf{37.58}} & \multirow{2}{*}{\textbf{0.23}} \\
    & WRN70-16 &  &  &  &  &  &  &  &  &  \\
    \addlinespace
    Debenedetti \cite{Debenedetti2023Light}& XCiT-L12 & 70.77 & 36.75 & 36.15 & \textbf{35.88} & \textbf{0.27} & 37.21 & 36.30 & \textbf{35.91} & \textbf{0.39} \\
    Debenedetti \cite{Debenedetti2023Light}& XCiT-M12 & 69.20 & 35.75 & 35.16 & \textbf{34.91} & \textbf{0.25} & 36.05 & 35.19 & \textbf{34.80} & \textbf{0.39} \\
    Gowal20 \cite{Gowal2020Uncovering}& WRN70-16 & 69.15 & 38.77 & 38.27 & \textbf{38.19} & \textbf{0.08} & 38.93 & 38.02 & \textbf{37.80} & \textbf{0.22} \\
    Pang \cite{Pang2022Robustness}& WRN70-16 & 65.56 & 34.14 & 33.77 & \textbf{33.61} & \textbf{0.16} & 34.22 & 33.82 & \textbf{33.64} & \textbf{0.18} \\
    Rebuffi \cite{Rebuffi2021Fixing}& WRN70-16 & 63.56 & 36.06 & 35.62 & \textbf{35.53} & \textbf{0.09} & 36.09 & 35.30 & \textbf{35.22} & \textbf{0.08} \\
    Wang \cite{Wang2023Better}& WRN70-16 & 75.22 & 43.84 & 43.55 & \textbf{43.47} & \textbf{0.08} & 43.83 & 43.63 & \textbf{43.32} & \textbf{0.31} \\
    Cui \cite{Cui2023Decoupled}& WRN28-10 & 73.85 & 40.28 & \textbf{39.93} & 39.96 & -0.03 & 40.29 & 40.13 & \textbf{39.78} & \textbf{0.35} \\
    Wang \cite{Wang2023Better}& WRN28-10 & 72.58 & 39.58 & 39.29 & \textbf{39.20} & \textbf{0.09} & 39.60 & 39.41 & \textbf{39.27} & \textbf{0.14} \\
    \addlinespace
    \cmidrule(lr){1-3}
    \multicolumn{3}{c}{ImageNet ($\varepsilon=4/255$)} &&&&&&&&\\
    \midrule
    Peng \cite{Peng2023Robust}& WRN101-2 & 73.10 & 50.92 & 50.40 & \textbf{50.20} & \textbf{0.20} & 51.32 & 50.36 & \textbf{50.06} & \textbf{0.30} \\
    Liu \cite{Liu2023Comprehensive}& Swin-B & 76.16 & 57.80 & 57.26 & \textbf{57.12} & \textbf{0.14} & 58.04 & 57.12 & \textbf{56.96} & \textbf{0.16} \\
    Liu & Swin-L & 78.92 & 61.34 & 60.84 & \textbf{60.74} & \textbf{0.10} & 61.76 & 60.78 & \textbf{60.56} & \textbf{0.22} \\
    Liu & ConvNeXt-L & 78.02 & 60.56 & 59.96 & \textbf{59.78} & \textbf{0.18} & 60.94 & 59.96 & \textbf{59.54} & \textbf{0.42} \\
    Liu & ConvNeXt-B & 76.70 & 58.06 & 57.44 & \textbf{57.18} & \textbf{0.26} & 58.56 & 57.26 & \textbf{57.06} & \textbf{0.20} \\ 
    \addlinespace
    \multirow{2}{*}{Singh \cite{Singh2023Revisiting}} & ViT-B+ & \multirow{2}{*}{76.30} & \multirow{2}{*}{56.74} & \multirow{2}{*}{56.04} & \multirow{2}{*}{\textbf{55.78}} & \multirow{2}{*}{\textbf{0.26}} & \multirow{2}{*}{57.30} & \multirow{2}{*}{56.22} & \multirow{2}{*}{\textbf{55.82}} & \multirow{2}{*}{\textbf{0.40}} \\
     & ConvStem &  &  &  &  &  &  &  &  &  \\
    \addlinespace
    \multirow{2}{*}{Singh} & ConvNeXt-B & \multirow{2}{*}{75.88} & \multirow{2}{*}{58.04} & \multirow{2}{*}{57.38} & \multirow{2}{*}{\textbf{57.18}} & \multirow{2}{*}{\textbf{0.20}} & \multirow{2}{*}{58.56} & \multirow{2}{*}{57.64} & \multirow{2}{*}{\textbf{57.48}} & \multirow{2}{*}{\textbf{0.16}} \\
     & +ConvStem &  &  &  &  &  &  &  &  &  \\
    \addlinespace
    \multirow{2}{*}{Singh} & ConvNeXt-S & \multirow{2}{*}{74.08} & \multirow{2}{*}{54.34} & \multirow{2}{*}{53.78} & \multirow{2}{*}{\textbf{53.56}} & \multirow{2}{*}{\textbf{0.22}} & \multirow{2}{*}{54.90} & \multirow{2}{*}{53.82} & \multirow{2}{*}{\textbf{53.38}} & \multirow{2}{*}{\textbf{0.44}} \\
    & +ConvStem &  &  &  &  &  &  &  &  &  \\
    \addlinespace
    \multirow{2}{*}{Singh} & ConvNeXt-T & \multirow{2}{*}{72.70} & \multirow{2}{*}{51.66} & \multirow{2}{*}{51.04} & \multirow{2}{*}{\textbf{50.70}} & \multirow{2}{*}{\textbf{0.34}} & \multirow{2}{*}{52.04} & \multirow{2}{*}{50.82} & \multirow{2}{*}{\textbf{50.56}} & \multirow{2}{*}{\textbf{0.26}} \\
     & +ConvStem &  &  &  &  &  &  &  &  &  \\
    \addlinespace
    \multirow{2}{*}{Singh} & ConvNeXt-L & \multirow{2}{*}{77.00} & \multirow{2}{*}{59.16} & \multirow{2}{*}{58.80} & \multirow{2}{*}{\textbf{58.70}} & \multirow{2}{*}{\textbf{0.10}} & \multirow{2}{*}{59.78} & \multirow{2}{*}{58.68} & \multirow{2}{*}{\textbf{58.50}} & \multirow{2}{*}{\textbf{0.18}} \\
     & +ConvStem &  &  &  &  &  &  &  &  &  \\
    \toprule
    \multirow{4}{*}{Summary} & \multicolumn{2}{r}{CIFAR-10 (\textbf{\#bold})} & 1 & 1 & \textbf{8} & & 1 & 0 & \textbf{9} & \\
    &\multicolumn{2}{r}{CIFAR-100 (\textbf{\#bold})} & 2 & 1 & \textbf{7} & & 1 & 0 & \textbf{9} & \\
    &\multicolumn{2}{r}{ImageNet (\textbf{\#bold})} & 0 & 0 & \textbf{10} & & 0 & 0 & \textbf{10} & \\
    \cmidrule(lr){2-11}
    &\multicolumn{2}{r}{\textbf{Total (\#bold)}} & 3 & 2 & \textbf{25} & & 2 & 0 & \textbf{28} & \\
    \bottomrule
    \end{tabular}
    }
\end{table}
\subsection{Experiments on ReACG}
\label{sec:experiments_main}
ReACG was compared with APGD and ACG.
This experiment used 30 models, including the top 10 for each dataset listed on the RobustBench leaderboard as of October 1, 2023.
All compared methods used the input point as the initial point, $N=100$ as the total number of iterations, and CW/DLR loss as the objective function.
\Cref{tab:full_cifar10_100} shows the robust accuracy obtained by each attack.
$\Delta$ represents the difference in robust accuracy between ACG and ReACG.
From the ``CW loss'' columns in \cref{tab:full_cifar10_100}, ReACG showed lower robust accuracy for 90\% of the 30 models than that in APGD and ACG.
With DLR loss, ReACG showed a higher attack performance than APGD and ACG for 93\% and 97\% of the 30 models, respectively.
ReACG showed lower robust accuracy than APGD and ACG by approximately 0.4 to 0.9\% and 0.1 to 0.4\%, respectively. Additionally, the difference in robust accuracy among APGD, ACG, and ReACG was higher with DLR loss than that with CW loss.
These improvements are sufficient considering the recent advances in adversarial attacks based on nonlinear optimization methods.
\begin{figure}[t]
    \centering
    \includegraphics[width=0.95\linewidth]{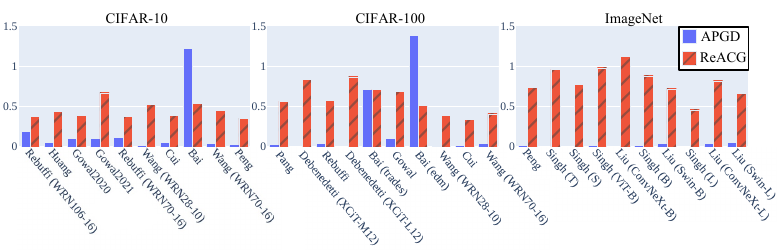}
    \caption{The percentage of images where APGD/ReACG found adversarial examples but ReACG/APGD did not.}
    \label{fig:comparison_only_success}
\end{figure}

\Cref{fig:comparison_only_success} shows the percentage of images where APGD/ReACG attacked successfully and ReACG/APGD attacked unsuccessfully.
APGD exhibited a lower robust accuracy than that of ReACG for the models proposed by Bai et al. \cite{Bai2023Improving}. However, \cref{fig:comparison_only_success} shows that ReACG successfully perturbs the images for which APGD fails. 
This result suggests that the ensemble of ReACG and APGD is likely to exhibit a high attack performance.

\subsection{Effect of rescaling}
\label{sec:analysis_rescaling}
\begin{figure}
  \centering
  \includegraphics[width=0.85\linewidth]{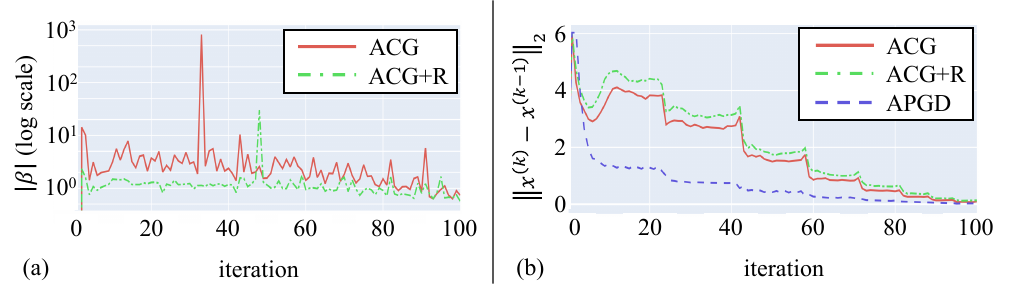}
  \caption{(a) Transition of $|\iter{\beta_{HS}}{k}|$.
  (b) Transition of the moving distance per iteration.
  }\label{fig:abs_beta_transition}
\end{figure}

\Cref{fig:abs_beta_transition}(a) and (b) show the transitions of $|\iter{\beta}{k}|$ and $\|\iter{\bm{x}}{k}-\iter{\bm{x}}{k-1}\|_2$, respectively. 
ACG+R adopts $\iter{\beta}{k}$-rescaling.
\Cref{fig:abs_beta_transition}(a) demonstrates that ACG+R has smaller $|\iter{\beta}{k}|$ than that of ACG, which indicates that our rescaling method reduces $|\iter{\beta}{k}|$.
Additionally, \cref{fig:abs_beta_transition}(b) shows that the search points of ACG+R move more than those of ACG. 
These results suggest that our rescaling method enhances the search efficiency by resolving the issue caused by a large $|\iter{\beta}{k}|$.

\subsection{Difference in CTC variation}
\label{sec:analysis_ctc_variation}
\begin{figure}
  \centering
  \includegraphics[width=0.85\linewidth]{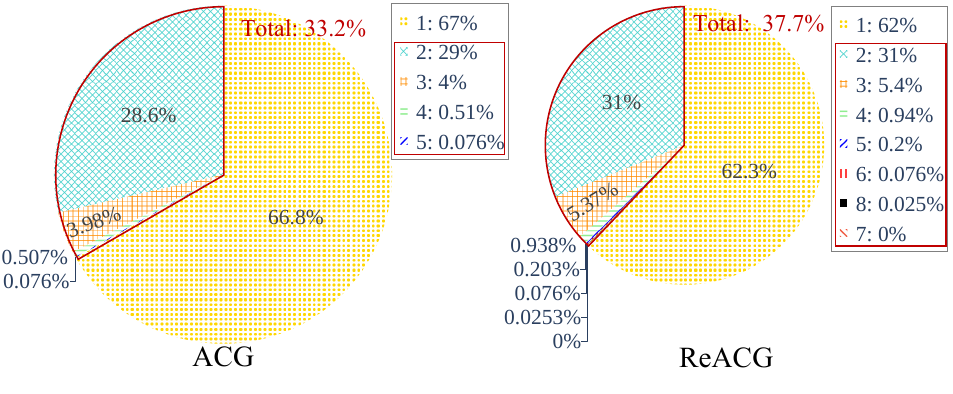}
  \caption{Comparison in CTC variation. Let $\iter{c}{k}=\arg\max_{i\neq c}f_i(\iter{\bm{x}}{k})$ be a CTC at $k$-th iterateion.
The number of CTCs appeared during an attack, \#CTC, is defined as $\textrm{\#CTC}:=|\{\iter{c}{i}\mid i=1,\ldots,N\}|$. In the legend, ``$K$: P\%'' means that the percentage of images for which \#CTC=$K$ is P\%.}\label{fig:ctc_variation}
\end{figure}

In \cref{sec:analysis_rescaling}, we discussed the effect of our rescaling method in the input space. This section highlights one reason for the high performance of ReACG by focusing on the difference in the output diversity of ACG and ReACG. 
Analysis of \cref{sec:analysis_rescaling}
implies that ReACG performs a more diversified search than that of ACG using appropriate step-size control in addition to rescaling $\iter{\beta}{k}$. The analysis in this section is based on CTC variation.
Let $\iter{c}{k}=\arg\max_{i\neq c}f_i(\iter{\bm{x}}{k})$ be the CTC at the $k$th iterateion.
The number of CTCs that appear during an attack \#CTC, is defined as $\textrm{\#CTC}:=|\{\iter{c}{i}\mid i=1,\ldots,N\}|$.
The left/right of \cref{fig:ctc_variation} show the percentage of images in which \#CTC=$K$ during an attack with ACG/ReACG.
\Cref{fig:ctc_variation} shows that ReACG has a smaller percentage of images where \#CTC=1 than that in ACG and a larger percentage of images where \#CTC$\geq2$. Additionally, ReACG has a larger maximum number of \#CTCs than that in ACG. These results suggest that ReACG enhances the output diversity compared to that with ACG.

\begin{table}
    \centering
    \caption{
    The results of the ablation study. 
    $\Delta$ is the difference in robust accuracy between ACG and ReACG. The lowest value for each model and loss is in bold.}
    \label{tab:results_ascg_cw}
    \scalebox{0.82}{
    \begin{tabular}{cccccccccc}
        \toprule
        \multicolumn{10}{c}{CW loss} \\
        \midrule
        Dataset & \multicolumn{3}{c}{CIFAR-10 ($\varepsilon$=8/255)} & \multicolumn{3}{c}{CIFAR-100 ($\varepsilon$=8/255)} & \multicolumn{3}{c}{ImageNet ($\varepsilon$=4/255)}\\
        \cmidrule(lr){1-1}
        \cmidrule(lr){2-4}
        \cmidrule(lr){5-7}
        \cmidrule(lr){8-10}
        Defense & Peng \cite{Peng2023Robust} & Wang \cite{Wang2023Better} & Bai \cite{Bai2023Improving} & Wang \cite{Wang2023Better} & Cui \cite{Cui2023Decoupled} & Wang \cite{Wang2023Better} & Liu \cite{Liu2023Comprehensive} & Liu \cite{Liu2023Comprehensive} & Singh \cite{Singh2023Revisiting} \\
        \midrule
        \multirow{2}{*}{Model} & WRN & WRN & RN-152+ & WRN & WRN & WRN & \multirow{2}{*}{Swin-L} & Conv- & ConvNeXt-L \\
         & 70-16 & 70-16 & WRN-70-16 & 70-16 & 28-10 & 28-10 &  & NeXT-L & +ConvStem \\
        \midrule
        clean & 93.27 & 93.25 & 95.23 & 75.22 & 73.85 & 72.58 & 78.92 & 78.02 & 77.00 \\
        \midrule
        APGD & 71.92 & 71.57 & \textbf{69.42} & 43.84 & 40.28 & 39.58 & 61.34 & 60.56 & 59.16 \\
        ACG & 71.66 & 71.25 & 69.78 & 43.55 & 39.93 & 39.29 & 60.84 & 59.96 & 58.80 \\
        ACG+R & 71.66 & 71.22 & 69.83 & 43.52 & 39.96 & 39.27 & \textbf{60.74} & 59.86 & 58.72\\
        ACG+T & 71.68 & 71.23 & 70.04 & \textbf{43.42} & \textbf{39.88} & 39.24 & 60.80 & 59.84 & 58.74 \\
        ReACG & \textbf{71.60} & \textbf{71.17} & 70.10 & 43.47 & 39.96 & \textbf{39.20} & \textbf{60.74} & \textbf{59.78} & \textbf{58.70} \\
        \midrule
        $\Delta ~(\uparrow)$ & 0.06 & 0.08 & -0.32 & 0.08 & -0.03 & 0.09 & 0.10 & 0.18 & 0.10 \\
        \bottomrule
        \addlinespace
        \multicolumn{10}{c}{DLR loss} \\
        \toprule
        APGD & 71.99 & 71.58 & \textbf{69.38} & 43.83 & 40.29 & 39.60 & 61.76 & 60.94 & 59.78 \\
        ACG & 71.53 & 71.32 & 69.86 & 43.63 & 40.13 & 39.41 & 60.78 & 59.96 & 58.68 \\
        ACG+R & 71.52 & 71.21 & 69.56 & 43.55 & 39.96 & 39.35 & 60.74 & 59.82 & 58.66\\
        ACG+T & 71.55 & 71.24 & 69.91 & 43.52 & 39.96 & \textbf{39.25} & 60.66 & 59.74 & 58.62 \\
        ReACG & \textbf{71.41} & \textbf{71.13} & 69.72 & \textbf{43.32} & \textbf{39.78} & 39.27 & \textbf{60.56} & \textbf{59.54} & \textbf{58.50} \\
        \midrule
        $\Delta ~(\uparrow)$ & 0.12 & 0.19 & 0.14 & 0.31 & 0.35 & 0.14 & 0.22 & 0.42 & 0.18 \\
        \bottomrule
    \end{tabular}
    }
\end{table}
\subsection{Ablation study}
\label{sec:experiments_ablation}
\subsubsection{Effects of rescaling and step size optimization}
\Cref{tab:results_ascg_cw} describes the robust accuracy for nine models, including the top three for each dataset listed in the RobustBench leaderboard.
This experiment used the same initial points, total number of iterations, and objective functions as those in \cref{sec:experiments_main}.
ACG+T is ACG that uses the parameters $p_1, q$ and $q_{min}$ selected in \cref{sec:experiments_tuning}. 

In \cref{tab:results_ascg_cw}, ACG+R and ACG+T show lower robust accuracies than that of ACG in many cases.
This result indicates that rescaling $\iter{\beta}{k}$ and step size optimization can improve attack performance.
Additionally, ACG+T showed the same or slightly lower robust accuracy than that of ACG+R, and ReACG showed a lower robust accuracy than that of ACG+R and ACG+T in several cases. 

\subsubsection{Relationship between the number of total iterations and robust accuracy}
\begin{figure}
    \centering
    \includegraphics[width=0.95\linewidth]{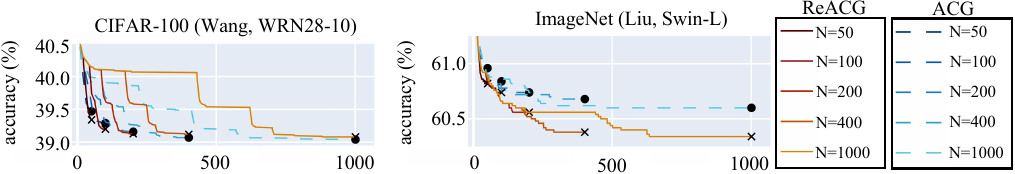}
    \caption{ACG vs. ReACG for different maximum numbers of iterations $N$.}
    \label{fig:several_iter}
\end{figure}

\Cref{fig:several_iter} shows the transition of the robust accuracy of ACG and ReACG with $N=50, 100, 200, 400,$ and $1000$. The dashed line in the figure represents the ACG, and the solid line represents the ReACG.
The left figure shows the typical results for CIFAR-10/100, and the right figure shows the typical results for ImageNet.
As shown in the figure on the left, the decrease in robust accuracy by ReACG was slower than that by ACG for CIFAR-10/100. At approximately $N \leq 200$, ReACG recorded a lower robust accuracy than that of ACG; however, at $N\geq 400$, ACG showed a lower robust accuracy than ReACG.
As shown in the figure on the right, the decrease in robust accuracy by ReACG was faster than that by ACG for ImageNet. Additionally, ReACG exhibited a lower robust accuracy than that of ACG for all $N$.
Similar trends were observed in the remaining models.

\subsubsection{Relationship between the number of restarts and robust accuracy}
\begin{figure}
    \centering
    \includegraphics[width=0.95\linewidth]{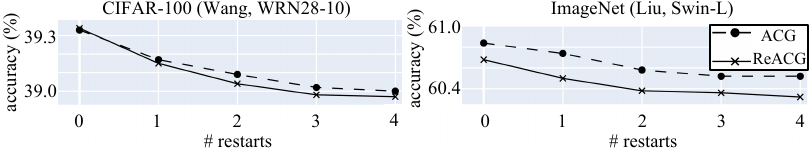}
    \caption{ACG vs. ReACG for different numbers of restarts.}
    \label{fig:multirestarts_cw_swin_l}
\end{figure}

This experiment investigated the effect of random restarts on the attack performance of ACG and ReACG. The initial points were sampled uniformly from a feasible region.
The left part of \cref{fig:multirestarts_cw_swin_l} shows the typical results for models trained on CIFAR-10/100, and the right part shows representative results for models trained on ImageNet.
\Cref{fig:multirestarts_cw_swin_l} shows that the robust accuracy decreases as the number of restarts increases. 
Additionally, ReACG showed a lower robust accuracy than that of ACG, even with multiple restarts. The same trend was observed in the other models.

\section{Conclusion}
We hypothesized that increasing the distance between two consecutive search points would enhance the output diversity, resulting in high attack performance.
We propose ReACG with an improved ACG search direction and step-size control to test our hypothesis.
ReACG automatically modifies $\iter{\beta}{k}$ when $|\iter{\beta}{k}|$ exceeds the average ratio of the gradient to conjugate gradient. Our analyses show that modifying $\iter{\beta}{k}$ and step size control increased the distance between two consecutive search points and CTC diversity, which led to the high attack performance of ReACG. 
ReACG showed better attack performance than the baseline methods APGD and ACG for 90\% of the 30 SOTA robust models trained on the three representative datasets.
The differences in the robust accuracy between APGD and ACG were approximately 0.4 to 0.9\% and 0.1 to 0.4\%, respectively. These are sufficiently large considering the recent advances in adversarial attacks based on nonlinear optimization methods.
Additionally, ReACG demonstrated particularly promising results for ImageNet models with a large number of classification classes. This result supports the claim that the output diversity contributes to high attack performance.
Increasing the distance between two consecutive search points to enhance the output diversity may be beneficial for developing new potent attacks.

\begin{credits}
\subsubsection{\ackname} 
This research project was supported by the Japan Science and Technology Agency (JST), Core Research of Evolutionary Science and Technology (CREST), Center of Innovation Science and Technology based Radical Innovation and Entrepreneurship Program (COI Program), JSPS KAKENHI Grant Numbers JP16H01707 and JP21H04599, Japan.

\subsubsection{\discintname}
The authors declare no competing interests relevant to the contents of this article. 
\end{credits}
\bibliographystyle{plainnat}
\bibliography{reference}
\end{document}

%% file: math_commands.tex
\usepackage{amsmath,amsfonts,amssymb,bm}

\def\eqref#1{equation~\ref{#1}}
\def\Eqref#1{Equation~\ref{#1}}

\def\1{\bm{1}}

\DeclareMathAlphabet{\mathsfit}{\encodingdefault}{\sfdefault}{m}{sl}
\SetMathAlphabet{\mathsfit}{bold}{\encodingdefault}{\sfdefault}{bx}{n}













%% file: main.bbl
\begin{thebibliography}{38}
\providecommand{\natexlab}[1]{#1}
\providecommand{\url}[1]{\texttt{#1}}
\expandafter\ifx\csname urlstyle\endcsname\relax
  \providecommand{\doi}[1]{doi: #1}\else
  \providecommand{\doi}{doi: \begingroup \urlstyle{rm}\Url}\fi

\bibitem[Addepalli et~al.(2022)Addepalli, Jain, and
  Radhakrishnan]{Addepalli2022Efficient}
Sravanti Addepalli, Samyak Jain, and Venkatesh~Babu Radhakrishnan.
\newblock Efficient and effective augmentation strategy for adversarial
  training.
\newblock In \emph{NeurIPS}, 2022.

\bibitem[Adjabi et~al.(2020)Adjabi, Ouahabi, Benzaoui, and
  Taleb-Ahmed]{adjabi2020past}
Insaf Adjabi, Abdeldjalil Ouahabi, Amir Benzaoui, and Abdelmalik Taleb-Ahmed.
\newblock Past, present, and future of face recognition: A review.
\newblock \emph{Electronics}, 9\penalty0 (8):\penalty0 1188, 2020.

\bibitem[Akiba et~al.(2019)Akiba, Sano, Yanase, Ohta, and Koyama]{optuna}
Takuya Akiba, Shotaro Sano, Toshihiko Yanase, Takeru Ohta, and Masanori Koyama.
\newblock Optuna: {A} next-generation hyperparameter optimization framework.
\newblock In \emph{SIGKDD}, 2019.

\bibitem[Andriushchenko and Flammarion(2020)]{Andriushchenko2020Understanding}
Maksym Andriushchenko and Nicolas Flammarion.
\newblock Understanding and improving fast adversarial training.
\newblock In \emph{NeurIPS}, 2020.

\bibitem[Bai et~al.(2023)Bai, Anderson, Kim, and Sojoudi]{Bai2023Improving}
Yatong Bai, Brendon~G Anderson, Aerin Kim, and Somayeh Sojoudi.
\newblock Improving the accuracy-robustness trade-off of classifiers via
  adaptive smoothing.
\newblock \emph{arXiv preprint arXiv:2301.12554}, 2023.

\bibitem[Carlini and Wagner(2018)]{8424625}
Nicholas Carlini and David Wagner.
\newblock Audio adversarial examples: Targeted attacks on speech-to-text.
\newblock In \emph{SPW}, 2018.

\bibitem[Carlini and Wagner(2017)]{CW2017}
Nicholas Carlini and David~A. Wagner.
\newblock Towards evaluating the robustness of neural networks.
\newblock In \emph{SP}, 2017.

\bibitem[Croce and Hein(2020)]{croce2020reliable}
Francesco Croce and Matthias Hein.
\newblock Reliable evaluation of adversarial robustness with an ensemble of
  diverse parameter-free attacks.
\newblock In \emph{ICML}, 2020.

\bibitem[Croce et~al.(2021)Croce, Andriushchenko, Sehwag, Debenedetti,
  Flammarion, Chiang, Mittal, and Hein]{croce2021robustbench}
Francesco Croce, Maksym Andriushchenko, Vikash Sehwag, Edoardo Debenedetti,
  Nicolas Flammarion, Mung Chiang, Prateek Mittal, and Matthias Hein.
\newblock Robustbench: a standardized adversarial robustness benchmark.
\newblock In \emph{NeurIPS}, 2021.

\bibitem[Cui et~al.(2023)Cui, Tian, Zhong, Qi, Yu, and Zhang]{Cui2023Decoupled}
Jiequan Cui, Zhuotao Tian, Zhisheng Zhong, Xiaojuan Qi, Bei Yu, and Hanwang
  Zhang.
\newblock Decoupled kullback-leibler divergence loss.
\newblock \emph{arXiv preprint arXiv:2305.13948}, 2023.

\bibitem[Debenedetti et~al.(2023)Debenedetti, Sehwag, and
  Mittal]{Debenedetti2023Light}
Edoardo Debenedetti, Vikash Sehwag, and Prateek Mittal.
\newblock A light recipe to train robust vision transformers.
\newblock In \emph{SaTML}, 2023.

\bibitem[Goodfellow et~al.(2015)Goodfellow, Shlens, and
  Szegedy]{goodfellow2014explaining}
Ian~J. Goodfellow, Jonathon Shlens, and Christian Szegedy.
\newblock Explaining and harnessing adversarial examples.
\newblock In \emph{ICLR}, 2015.

\bibitem[Gowal et~al.(2019)Gowal, Uesato, Qin, Huang, Mann, and Kohli]{MTPGD}
Sven Gowal, Jonathan Uesato, Chongli Qin, Po{-}Sen Huang, Timothy~A. Mann, and
  Pushmeet Kohli.
\newblock An alternative surrogate loss for pgd-based adversarial testing.
\newblock \emph{CoRR}, abs/1910.09338, 2019.

\bibitem[Gowal et~al.(2020)Gowal, Qin, Uesato, Mann, and
  Kohli]{Gowal2020Uncovering}
Sven Gowal, Chongli Qin, Jonathan Uesato, Timothy~A. Mann, and Pushmeet Kohli.
\newblock Uncovering the limits of adversarial training against norm-bounded
  adversarial examples.
\newblock \emph{CoRR}, abs/2010.03593, 2020.

\bibitem[Gowal et~al.(2021)Gowal, Rebuffi, Wiles, Stimberg, Calian, and
  Mann]{Gowal2021Improving}
Sven Gowal, Sylvestre{-}Alvise Rebuffi, Olivia Wiles, Florian Stimberg,
  Dan~Andrei Calian, and Timothy~A. Mann.
\newblock Improving robustness using generated data.
\newblock In \emph{NeurIPS}, 2021.

\bibitem[Gupta et~al.(2021)Gupta, Anpalagan, Guan, and Khwaja]{gupta2021deep}
Abhishek Gupta, Alagan Anpalagan, Ling Guan, and Ahmed~Shaharyar Khwaja.
\newblock Deep learning for object detection and scene perception in
  self-driving cars: Survey, challenges, and open issues.
\newblock \emph{Array}, 10:\penalty0 100057, 2021.

\bibitem[Huang et~al.(2022)Huang, Lu, Deb, and Boddeti]{Huang2022Revisiting}
Shihua Huang, Zhichao Lu, Kalyanmoy Deb, and Vishnu~Naresh Boddeti.
\newblock Revisiting residual networks for adversarial robustness: An
  architectural perspective.
\newblock \emph{arXiv preprint arXiv:2212.11005}, 2022.

\bibitem[Krizhevsky et~al.(2009)Krizhevsky, Hinton,
  et~al.]{Krizhevsky09learningmultiple}
Alex Krizhevsky, Geoffrey Hinton, et~al.
\newblock Learning multiple layers of features from tiny images.
\newblock 2009.

\bibitem[Li et~al.(2020)Li, Deng, Li, Yan, Gao, and Huang]{Li_2020_CVPR}
Maosen Li, Cheng Deng, Tengjiao Li, Junchi Yan, Xinbo Gao, and Heng Huang.
\newblock Towards transferable targeted attack.
\newblock In \emph{CVPR}, 2020.

\bibitem[Lin et~al.(2022)Lin, Lucas, Bauer, Reiter, and Sharif]{CGD2022}
Weiran Lin, Keane Lucas, Lujo Bauer, Michael~K. Reiter, and Mahmood Sharif.
\newblock Constrained gradient descent: A powerful and principled evasion
  attack against neural networks.
\newblock In \emph{ICML}, 2022.

\bibitem[Liu et~al.(2023{\natexlab{a}})Liu, Dong, Xiang, Yang, Su, Zhu, Chen,
  He, Xue, and Zheng]{Liu2023Comprehensive}
Chang Liu, Yinpeng Dong, Wenzhao Xiang, Xiao Yang, Hang Su, Jun Zhu, Yuefeng
  Chen, Yuan He, Hui Xue, and Shibao Zheng.
\newblock A comprehensive study on robustness of image classification models:
  Benchmarking and rethinking.
\newblock \emph{arXiv preprint arXiv:2302.14301}, 2023{\natexlab{a}}.

\bibitem[Liu et~al.(2022{\natexlab{a}})Liu, Lin, Yang, Ng, Divakaran, and
  Dong]{liu2022inferring}
Ruofan Liu, Yun Lin, Xianglin Yang, Siang~Hwee Ng, Dinil~Mon Divakaran, and
  Jin~Song Dong.
\newblock Inferring phishing intention via webpage appearance and dynamics: A
  deep vision based approach.
\newblock In \emph{USENIX Security}, 2022{\natexlab{a}}.

\bibitem[Liu et~al.(2023{\natexlab{b}})Liu, Peng, and Tang]{AutoAE23}
Shengcai Liu, Fu~Peng, and Ke~Tang.
\newblock Reliable robustness evaluation via automatically constructed attack
  ensembles.
\newblock In \emph{AAAI}, 2023{\natexlab{b}}.

\bibitem[Liu et~al.(2022{\natexlab{b}})Liu, Cheng, Gao, Liu, Zhang, and
  Song]{Ye2022Practical}
Ye~Liu, Yaya Cheng, Lianli Gao, Xianglong Liu, Qilong Zhang, and Jingkuan Song.
\newblock Practical evaluation of adversarial robustness via adaptive auto
  attack.
\newblock In \emph{CVPR}, 2022{\natexlab{b}}.

\bibitem[Madry et~al.(2018)Madry, Makelov, Schmidt, Tsipras, and
  Vladu]{madry2018towards}
Aleksander Madry, Aleksandar Makelov, Ludwig Schmidt, Dimitris Tsipras, and
  Adrian Vladu.
\newblock Towards deep learning models resistant to adversarial attacks.
\newblock In \emph{ICLR}, 2018.

\bibitem[Mao et~al.(2021)Mao, Chen, Wang, Su, He, and Xue]{CAA2021}
Xiaofeng Mao, Yuefeng Chen, Shuhui Wang, Hang Su, Yuan He, and Hui Xue.
\newblock Composite adversarial attacks.
\newblock In \emph{AAAI}, 2021.

\bibitem[Pang et~al.(2022)Pang, Lin, Yang, Zhu, and Yan]{Pang2022Robustness}
Tianyu Pang, Min Lin, Xiao Yang, Jun Zhu, and Shuicheng Yan.
\newblock Robustness and accuracy could be reconcilable by (proper) definition.
\newblock In \emph{ICML}, 2022.

\bibitem[Peng et~al.(2023)Peng, Xu, Cornelius, Hull, Li, Duggal, Phute, Martin,
  and Chau]{Peng2023Robust}
ShengYun Peng, Weilin Xu, Cory Cornelius, Matthew Hull, Kevin Li, Rahul Duggal,
  Mansi Phute, Jason Martin, and Duen~Horng Chau.
\newblock Robust principles: Architectural design principles for adversarially
  robust cnns.
\newblock In \emph{BMVC}, 2023.

\bibitem[Rebuffi et~al.(2021)Rebuffi, Gowal, Calian, Stimberg, Wiles, and
  Mann]{Rebuffi2021Fixing}
Sylvestre{-}Alvise Rebuffi, Sven Gowal, Dan~Andrei Calian, Florian Stimberg,
  Olivia Wiles, and Timothy~A. Mann.
\newblock Data augmentation can improve robustness.
\newblock In \emph{NeurIPS}, 2021.

\bibitem[Russakovsky et~al.(2015)Russakovsky, Deng, Su, Krause, Satheesh, Ma,
  Huang, Karpathy, Khosla, Bernstein, Berg, and Fei-Fei]{ILSVRC15}
Olga Russakovsky, Jia Deng, Hao Su, Jonathan Krause, Sanjeev Satheesh, Sean Ma,
  Zhiheng Huang, Andrej Karpathy, Aditya Khosla, Michael Bernstein,
  Alexander~C. Berg, and Li~Fei-Fei.
\newblock {ImageNet Large Scale Visual Recognition Challenge}.
\newblock \emph{IJCV}, 115\penalty0 (3):\penalty0 211--252, 2015.

\bibitem[Sehwag et~al.(2022)Sehwag, Mahloujifar, Handina, Dai, Xiang, Chiang,
  and Mittal]{Sehwag2021Proxy}
Vikash Sehwag, Saeed Mahloujifar, Tinashe Handina, Sihui Dai, Chong Xiang, Mung
  Chiang, and Prateek Mittal.
\newblock Robust learning meets generative models: Can proxy distributions
  improve adversarial robustness?
\newblock In \emph{ICLR}, 2022.

\bibitem[Singh et~al.(2023)Singh, Croce, and Hein]{Singh2023Revisiting}
Naman~D Singh, Francesco Croce, and Matthias Hein.
\newblock Revisiting adversarial training for imagenet: Architectures, training
  and generalization across threat models.
\newblock \emph{arXiv preprint arXiv:2303.01870}, 2023.

\bibitem[Szegedy et~al.(2014)Szegedy, Zaremba, Sutskever, Bruna, Erhan,
  Goodfellow, and Fergus]{szegedy2013intriguing}
Christian Szegedy, Wojciech Zaremba, Ilya Sutskever, Joan Bruna, Dumitru Erhan,
  Ian~J. Goodfellow, and Rob Fergus.
\newblock Intriguing properties of neural networks.
\newblock In \emph{ICLR}, 2014.

\bibitem[Tashiro et~al.(2020)Tashiro, Song, and Ermon]{tashiro2020ods}
Yusuke Tashiro, Yang Song, and Stefano Ermon.
\newblock Diversity can be transferred: Output diversification for white- and
  black-box attacks.
\newblock In \emph{NeurIPS}, 2020.

\bibitem[Wang et~al.(2023)Wang, Pang, Du, Lin, Liu, and Yan]{Wang2023Better}
Zekai Wang, Tianyu Pang, Chao Du, Min Lin, Weiwei Liu, and Shuicheng Yan.
\newblock Better diffusion models further improve adversarial training.
\newblock In \emph{ICML}, 2023.

\bibitem[Wong et~al.(2020)Wong, Rice, and Kolter]{Wong2020Fast}
Eric Wong, Leslie Rice, and J.~Zico Kolter.
\newblock Fast is better than free: Revisiting adversarial training.
\newblock In \emph{ICLR}, 2020.

\bibitem[Yamamura et~al.(2022)Yamamura, Sato, Tateiwa, Hata, Mitsutake, Oe,
  Ishikura, and Fujisawa]{Yamamura2022}
Keiichiro Yamamura, Haruki Sato, Nariaki Tateiwa, Nozomi Hata, Toru Mitsutake,
  Issa Oe, Hiroki Ishikura, and Katsuki Fujisawa.
\newblock Diversified adversarial attacks based on conjugate gradient method.
\newblock In \emph{ICML}, 2022.

\bibitem[Yao et~al.(2020)Yao, He, Han, and Zhou]{yao2020miss}
Qingsong Yao, Zecheng He, Hu~Han, and S~Kevin Zhou.
\newblock Miss the point: Targeted adversarial attack on multiple landmark
  detection.
\newblock In \emph{MICCAI}, 2020.

\end{thebibliography}
